\documentclass[runningheads]{llncs}

\usepackage[utf8]{inputenc}
\usepackage{cite}
\usepackage{subcaption}

\usepackage{tikz}
\usepackage{tkz-euclide}
\usepackage{pgfplots}  
\usepackage{xspace}
\usepgfplotslibrary{polar} 
\usepgflibrary{shapes.geometric}  
\usetikzlibrary{calc}  
\usetikzlibrary{quotes,angles}
\usepackage{amssymb}

\newcommand{\deltax}{\Delta x\xspace}
\newcommand{\deltay}{\Delta y\xspace}
\newcommand{\deltaz}{\Delta z\xspace}

\newcommand{\expectederrortwod}{\mathcal{\epsilon}_{2D}}

    \title{Effect of Timing Error: A Case Study of Navigation Camera}
\author{Sandeep S. Kulkarni\inst{1} \and Sanjay M Joshi\inst{2}}
\institute{\email{sandeep@msu.edu}, Michigan State University  \and Globus Medical}
\date{}

\pgfplotsset{my style/.append style={axis x line=middle, axis y line=  
middle, xlabel={$x$}, ylabel={$y$}, axis equal }} 

\begin{document}

\maketitle

\vspace*{-5mm}

\begin{abstract}
    We focus on the problem of timing  errors in navigation camera as a case study in a broader problem of the effect of a timing error in cyber-physical systems. These systems rely on the requirement that certain things happen at the same time or certain things happen periodically at some period $T$. However, as these systems get more complex, timing errors can occur between the components thereby violating the assumption about events being simultaneous (or periodic). 
    
    We consider the problem of a surgical navigation system where optical markers detected in the 2D pictures taken by two cameras are used to localize the markers in 3D space. A predefined array of such markers, known as a reference element, is used to navigate the corresponding CAD model of a surgical instrument on patient's images. The cameras rely on the assumption that the pictures from both cameras are taken exactly at the same time. If a timing error occurs then the instrument may have moved between the pictures. We find that, depending upon the location of the instrument, this can lead to a substantial error in the localization of the instrument. Specifically, we find that if the actual movement is $\delta$ then the observed movement may be as high as $5\delta$ in the operating range of the camera. Furthermore, we also identify potential issues that could affect the error in case there are changes to the camera system or to the operating range. 
\end{abstract}

\section{Introduction}
\label{sec:intro}
\vspace*{-3mm}


This work aims to evaluate the effect of timing errors in cyber-physical systems. 
Specifically, many cyber-physical systems expect certain things to happen \textit{simultaneously} or \textit{periodically} with period $T$. However, if the events are happening under two different subsystems, then there may be a timing error that causes the events to happen an interval $\epsilon$ apart. There are several causes for such an error. One possibility is a clock drift between the subsystems. Yet another reason could be non-deterministic execution between the subsystems, where one subsystem ends up taking the action before the other.

We focus on the case study in surgical navigation \cite{ndivega}. Such systems consist of two cameras at a known position with respect to each other. Each camera takes a picture of the surgical field upon a trigger, in infrared light. Specially coated optical markers, typically spheres, glow brightly under infrared light, while the rest of the field appears dark. These spheres appear as bright circles in captured 2D images. Using image processing techniques, the center of these circles is estimated and is used as a point estimate of projection of the marker on a virtual plane. Theoretically, the marker may lie anywhere on the line connecting the camera to the detected point on a virtual plane. Similarly, processing a second picture taken by another camera some distance away yields another line. The intersection of these two lines yields the estimated 3D position of the optical marker (cf. Figure \ref{fig:cameradiagram}).

A predefined geometrical array of 3 or more markers creates a reference element. By matching detected positions of multiple markers in the surgical field, the 3D position and pose of the reference element is estimated in the coordinate system of the camera. To register patient's anatomical images in the camera space, a registration fixture that is `opaque' to anatomical imaging technique (e.g., metal ball-bearings are opaque in X-rays and CT) is used with a known reference element. The registration fixture is attached to the patient before taking the anatomical scan. Using pattern recognition, the position of the registration fixture is detected in the anatomical scan. When the camera detects the reference element of the registration fixture, the position and pose of the patient's anatomical images are estimated in camera space. The surgeon can then `navigate' an instrument attached to a reference element since the CAD model of the instrument and reference element can be overlaid on the patient's registered anatomical images. 
The accuracy of marker location estimation is typically within 0.03cm from its actual location.
Such navigation techniques significantly reduce multiple x-rays in minimally invasive surgeries, where the tip of an instrument is not visible directly.

\begin{figure*}[t]
\vspace*{-15mm}
    \subfloat[]{
        \includegraphics[width=0.55\textwidth]{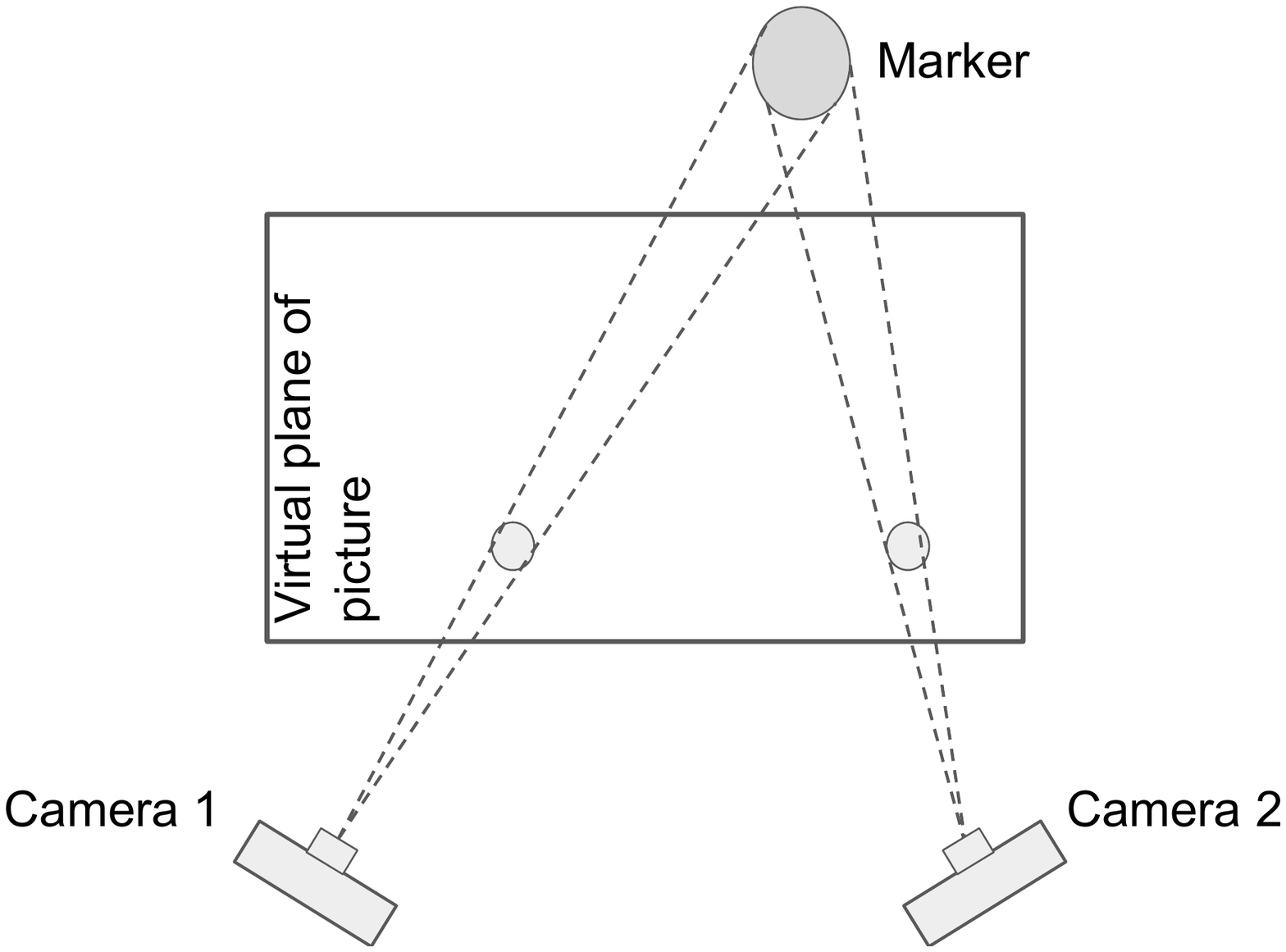}
        \vspace*{-5mm}
        \label{fig:nav_camera}}
    \hspace*{-20mm}
    \subfloat[]{
        \includegraphics[width=0.55\textwidth]{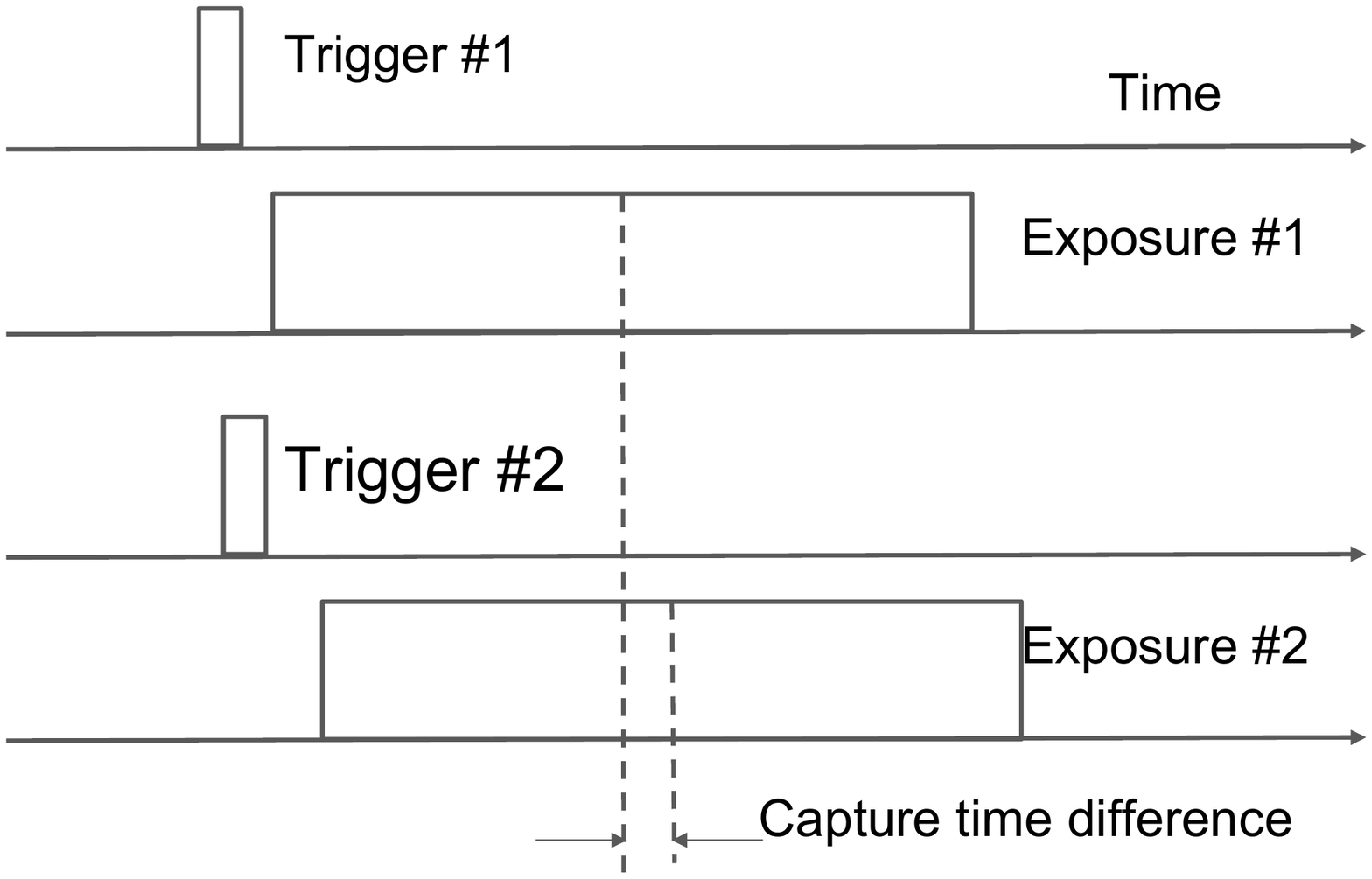}
        \vspace*{-5mm}
        \label{fig:timing_diff}}
        \caption{(a) Illustration (not to scale) of how two cameras capture a slightly different view of an optical marker, (b) Illustration (not to scale) of how slight differences in two cameras in arrival and processing of trigger, camera exposure window start, and camera exposure window width can lead to a difference in timing of picture}
        \vspace*{-5mm}

\end{figure*}

This technique relies on the assumption that the pictures taken by the two cameras are simultaneous, i.e., the instrument under consideration has not moved in between. The cameras typically employ charge-coupled devices (CCDs) to take digital pictures. Each CCD's exposure is started based on a trigger from the main processor. Each CCD's exposure window stays on for a predefined duration of time. The capture time of the picture is considered as the mid-point of this window. Any mismatches in timings or exposure can result in a difference in capture time as shown in Figure \ref{fig:timing_diff}. A capture time error could lead to a violation of the assumption of simultaneous capture time. Thus, a natural question is how such an error will impact the location identified by the camera. In this paper, we investigate the effect of such a timing error on the location of the marker identified by the system.

\textbf{Organization of the paper. }
We proceed as follows: First, in Section \ref{sec:navigationcamera2D}, we present the model for the navigation camera system for two dimensions.  Section \ref{sec:2Dproblem} analyzes the effect of timing error in 2D systems. In Section \ref{sec:2Dexperiments}, we present our simulation results of timing error in the 2D navigation camera system.
In Section \ref{sec:3Dproblem}, we extend the problem to three dimensions and analyze it in Section \ref{sec:3DAnalysis}. %
Simulation for 3D is performed in Section \ref{sec:3Dsimulation}. Finally, we discuss related work in Section \ref{sec:related} and make concluding remarks and discuss future work in Section \ref{sec:concl}.


\section{Modeling 2D Navigation Camera System}
\label{sec:navigationcamera2D}
\vspace*{-3mm}

\subsection{Model in the Absence of Timing Errors}
\begin{figure}[ht]
\vspace*{-7mm}
\centering    
\begin{tikzpicture}  

\draw[<->, gray, thick] (-4, 0) -- (4, 0);
\draw[->, gray, thick] (0, 0) -- (0, 3.2);

\draw[->, gray, thick] (-2, 0) -- (-2, 2);
\draw[->, gray, thick] (2, 0) -- (2, 2);

\draw[gray, thick] (0, 0) -- (0, 3.2);
\draw[blue, thick] (-2, 0) -- (3, 3);
\draw[blue, thick] (2, 0) -- (3, 3);
\filldraw[black] (3,3) circle (2pt) node[anchor=west] {C1 (x,y)};

\draw[<->, red,  thick] (-2, -0.2) -- (0, -0.2);

\draw (-2, -0.4) node {A};
\draw (2, -0.4) node {B};
\draw (0, -0.4) node {O};
\draw (4, -0.2) node {X axis};
\draw (0, 2.8) node {Y axis};


\draw (-1, -0.2) node {d};

\draw[<->, red,  thick] (2, -0.2) -- (0, -0.2);
\draw (1, -0.2) node {d};

\draw (1, 2) node {$r_1$};

\draw (2.3, 1.5) node {$r_2$};

\draw (-1.7, 0.45) node {$\beta_1$};
\draw (2.1, 0.45) node {$\beta_2$};

\draw (-1.5, 0.15) node {$\alpha_1$};
\draw (-1.3, 0) arc [
        start angle=0,
        end angle=5,
        x radius=5,
        y radius =5
    ] ;

\draw (2.5, 0.15) node {$\alpha_2$};
\draw (2.7, 0) arc [
        start angle=0,
        end angle=50,
        x radius=1,
        y radius =1
    ] ;

\end{tikzpicture}  
    \caption{Camera Diagram}
    \label{fig:cameradiagram}
\vspace*{-3mm}
\end{figure}
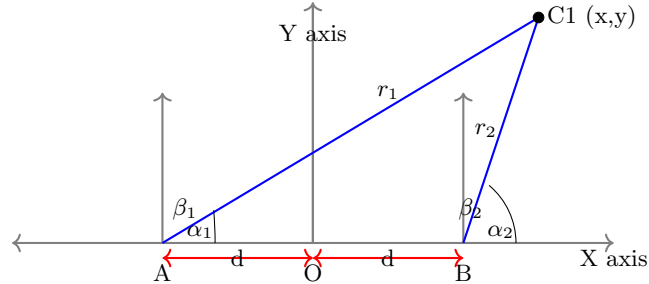

Based on the discussion in Section  \ref{sec:intro}, a navigation camera system has two cameras at points A and B, as shown in Figure \ref{fig:cameradiagram}, $2d$ distance apart. The camera coordinate system is defined as follows: The fixed rod connecting the two points is the $X$ axis. The center of the cameras is the origin of the coordinate system. The positive direction of the $Y$ axis is in the front of the camera, perpendicular to the $X$ axis. The $Z$ axis points up from the page to create a right-handed coordinate system.

Let the center of a marker $C1$ in the camera field be at location $(x, y)$. 
Let $\alpha_1$ and $\alpha_2$ be the angles with respect to X axis formed at $A$ and $B$. Let $\beta_1, \beta_2$ be the angles reported with $Y$ axis. Subsequently, the camera system software is responsible for computing the location $(x,y)$ based on  $\alpha_1, \alpha_2, \beta_1$ and $\beta_2$.


\subsection{Effect of Timing Error in Navigation Camera}

In this section, we identify the problem caused by timing error in navigation camera setting. 
We consider the case where the instrument held by the surgeon is at location $(x,y)$, when Camera 1 takes the picture of point $C1$. However, due to timing errors,  Camera 2 takes the picture slightly earlier when point $C1$ was at location $(x1, y1)$, where $x1=x+\deltax$, and $y1 =y+\deltay$. We expect $\deltax$ and $\deltay$ to be small values.

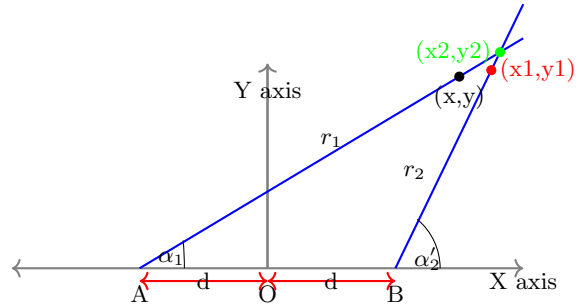
\begin{figure}[ht]
\centering    
\begin{tikzpicture}  [scale=0.85]

\draw[<->, gray, thick] (-4, 0) -- (4, 0);
\draw[->, gray, thick] (0, 0) -- (0, 3.2);

\draw[gray, thick] (0, 0) -- (0, 3.2);
\draw[blue, thick] (-2, 0) -- (3, 3);
\draw[blue, thick] (3, 3) -- (4, 3.6);

\draw[blue, thick] (2, 0) -- (3.5, 3.1);
\draw[blue, thick] (3.5, 3.1) -- (4, 4.13333);

\filldraw[black] (3,3) circle (2pt) node[anchor=north] {(x,y)};

\filldraw[red] (3.5,3.1) circle (2pt) node[anchor=west] {(x1,y1)};

\filldraw[green] (3.636,3.382) circle (2pt) node[anchor=east] {(x2,y2)};

\draw[<->, red,  thick] (-2, -0.2) -- (0, -0.2);

\draw (-2, -0.4) node {A};
\draw (2, -0.4) node {B};
\draw (0, -0.4) node {O};
\draw (4, -0.2) node {X axis};
\draw (0, 2.8) node {Y axis};


\draw (-1, -0.2) node {d};

\draw[<->, red,  thick] (2, -0.2) -- (0, -0.2);
\draw (1, -0.2) node {d};

\draw (1, 2) node {$r_1$};

\draw (2.3, 1.5) node {$r_2$};

\draw (-1.5, 0.15) node {$\alpha_1$};
\draw (-1.3, 0) arc [
        start angle=0,
        end angle=5,
        x radius=5,
        y radius =5
    ] ;

\draw (2.5, 0.15) node {$\alpha_2'$};
\draw (2.7, 0) arc [
        start angle=0,
        end angle=50,
        x radius=1,
        y radius =1
    ] ;

\end{tikzpicture}  

    \caption{Camera Diagram}
    \label{fig:cameraerror}
    \vspace*{-5mm}
\end{figure}

The value of $\deltax$ and $\deltay$ would depend upon the timing error and the movement rate. For example, if the timing error between cameras were 1 ms and the rate of the movement is 0.1 m/s then the total movement ($\sqrt{\deltax^2 + \deltay^2}) $ due to the timing error would be 0.1 mm. If timing error were 0.1 ms and the rate of movement is 0.1m/s then the movement would be 0.01mm 
(As a reference, in \cite{ndivega}, the speed is expected to be under $0.4m/s$.)

When a timing error occurs, the angles reported by the cameras would be as shown in Figure \ref{fig:cameraerror}.
It follows that Camera 1 will use location $(x, y)$ and report the angles. 
However, Camera 2 will use angles based on the point $(x1, y1)$. 
The actual location will be then determined by the angles $\alpha_1$ and $\beta_1$ reported by Camera 1 and $\alpha_2'$ and $\beta_2'$ by Camera 2. This is shown by location $(x2, y2)$ in Figure \ref{fig:cameraerror}. 

From the above discussion, it follows that if the timing error causes the movement from $(x,y)$ to $(x1, y1)$ then the computed error would be $(x, y)$ to $(x2, y2)$. In other words, the observed error would be $(x2-x, y2-y)$. Our goal is to compute this value based on the values of $x,y,\deltax$ and $\deltay$. 

\textbf{Remark. } Note that the decision to compute the error with respect to location $(x,y)$ is arbitrary; the results would be similar if we had computed the error with respect to $(x1, y1)$ due to symmetry where Camera 1 detects $C1$ and location $(x1,y1)$ and Camera 2 detects it at location $(x, y)$.
{For simplicity of analysis, we use $(x,y)$ as reference point in the analysis. We use the midpoint of $(x, y)$ and $(x1, y1)$ in the simulation.}
 
\vspace*{-3mm}
\section{Analysis of 2D Camera Problem}
\label{sec:2Dproblem}
\vspace*{-3mm}

In this section, we analyze the 2D camera problem in detail. First, in Section \ref{sec:computelocations}, we identify how the location can be determined based on the angles reported by the cameras. Subsequently, in Section \ref{sec:computeangles}, we solve the reverse problem to determine angles reported by cameras. This analysis would be useful to determine the angles reported by Camera 2 due to timing error. In Section \ref{sec:2dsymbolic}, we symbolically/analytically compute the localization error due to timing error. 

\subsection{Identifying Location of Point Based on Camera Angles}
\label{sec:computelocations}

The goal of this section is to discuss the relevant formula to determine how the angles reported by the cameras can be used to determine the location (x, y). From Figure \ref{fig:cameradiagram}, we have 

\vspace*{-3mm}
\begin{equation}
x \ \ =\ \  r_1 * \cos(\alpha_1) - d \ \ =\ \  
 r_2 * \cos(\alpha_2) + d \label{eq:xval}
\end{equation} 
\vspace*{-3mm}

\begin{equation}
  y \ \ = \ \ r_1 * \cos(\beta_1) \  \ = \ \ r_2 * \cos(\beta_2)  \label{eq:yval}
\end{equation}
\vspace*{-3mm}

From Equation \ref{eq:yval}, we have $r_1 = \frac{y}{\cos \beta_1}$. And, $r_2 = \frac{y}{\cos \beta_2}$. Substituting this in Equation \ref{eq:xval}, we have 

\begin{equation}
x \ \ =\ \  \frac{y}{\cos \beta_1} * \cos(\alpha_1) - d \ \ =\ \  
 \frac{y}{\cos \beta_2} * \cos(\alpha_2) + d \label{eq:xy}
\end{equation}

$$ \therefore  \frac{y}{\cos \beta_1} * \cos(\alpha_1) - \frac{y}{\cos \beta_2} * \cos(\alpha_2) \ \ =\ \  
  2d 
$$

$ \therefore y* \cos (\beta_2) * \cos(\alpha_1) - y * \cos (\beta_1) * \cos(\alpha_2) =  
  2d * \cos (\beta_1) * \cos(\beta_2)$

$$ \therefore y = \frac{2d * \cos (\beta_1) * \cos(\beta_2)}{\cos(\beta_2)*\cos(\alpha_1) - \cos(\beta_1)*\cos(\alpha_2)}
$$

Furthermore, $\tan(\alpha_1) = \frac{y}{x+d}$. Hence, $x+d = \frac{y}{\tan(\alpha_1)}$, i.e., 
\begin{small}
$$
x+d = \frac{2d * \cos (\beta_1) * \cos(\beta_2)}{\cos(\beta_2)*\cos(\alpha_1) - \cos(\beta_1)*\cos(\alpha_2)} * \frac{\cos(\alpha_1)}{\sin(\alpha_1)}
$$

$$ \therefore
x = \frac{2d * \cos (\beta_1) * \cos(\beta_2)}{\cos(\beta_2)*\cos(\alpha_1) - \cos(\beta_1)*\cos(\alpha_2)} * \frac{\cos(\alpha_1)}{\sin(\alpha_1)} - d
$$
%
\end{small}
\subsection{Identifying Camera Angles Based on Location}
\label{sec:computeangles}

In this section, we focus on identifying the angles reported by the camera when given a location $(x, y)$. In other words, the analysis focuses on the inverse of the problem in Section \ref{sec:computelocations}.
The purpose of this analysis is to identify what angles the camera would report if it observed the location at $(x1, y1)$ due to a timing error. 
From Figure \ref{fig:cameradiagram}, we have 
$\tan(\alpha_1) = y/(x+d)$.
$$
\therefore \alpha_1 = \arctan\left( \frac{y}{x+d} \right) \quad
, \quad
\alpha_2 = \arctan\left( \frac{y}{x-d} \right)
$$






Additionally, in the 2D model, $\beta_1 = \frac{\pi}{2} - \alpha_1$ and $\beta_2 = \frac{\pi}{2} - \alpha_2$.


\subsection{Symbolic Analysis of Localization Error due to Timing Error}
\label{sec:2dsymbolic}

In this section, we symbolically analyze the localization error due to timing error.
Specifically, we consider the analysis from Sections \ref{sec:computelocations} and \ref{sec:computeangles} symbolically to determine the error caused by the fact that the pictures were taken at different times. 
As mentioned earlier, we begin with a location $(x, y)$. We use the angles $\alpha_1$, $\alpha_2$, $\beta_1$, and $\beta_2$ identified by Camera 1 and 2. If the object is moving with a velocity
$(v_x, v_y)$ and the second camera has a timing error of $\Delta t$, then the error introduced in location is $(\deltax, \deltay) = (v_x\Delta t, v_y\Delta t)$. Subsequently, we use $x1$ (i.e., $(x+\deltax)$) and $y1$ (i.e., $(y+\deltay)$) to determine $\alpha_2'$ and $\beta_2'$. Finally, we use $\alpha_1$, $\beta_1$, $\alpha_2'$ and $\beta_2'$ to determine $x2$ and $y2$. Thus, the error would be $(x2-x, y2-y)$

From Section \ref{sec:computelocations}, we have 

$$ y2 = \frac{2d * \cos (\beta_1) * \cos(\beta_2')}{\cos(\beta_2')*\cos(\alpha_1) - \cos(\beta_1)*\cos(\alpha_2')}
$$

Using the fact that $\cos (\beta_1) = \sin(\alpha_1)$ and  $\cos (\beta_2') = \sin(\alpha_2')$, and dividing numerator and denominator by  $\cos (\beta_1) * \cos(\beta_2')$, we have

$$
\therefore y2 = 2d*\frac{1}{\frac{\cos(\alpha_1)}{\sin(\alpha_1)} - \frac{\cos(\alpha_2')}{\sin(\alpha_2')}}
= 2d*\frac{1}{\frac{1}{\tan(\alpha_1)} - \frac{1}{\tan(\alpha_2')}}
$$

$$
\therefore y2 = 2d*\frac{\tan(\alpha_1) \tan(\alpha_2')}{\tan(\alpha_2') - \tan(\alpha_1)}
$$

Now, we use the formula for $\alpha_1$ from Section \ref{sec:computeangles}. Note that due to error in the time the picture was taken, $\alpha_2$ would be based on the location $(x1,y1)=(x+\deltax, y+\deltay)$ instead of $(x,y)$

$$
\therefore y2 = 2d*\frac{\frac{y}{x+d}\cdot\frac{y+\deltay}{x+\deltax-d}}{\frac{y+\deltay}{x+\deltax-d}- \frac{y}{x+d}}
$$

Multiplying by $(x+d)(x+\deltax-d)$ on both the numerator and denominator, we have
$$
 y2 = 2d*\frac{y^2+y\deltay}{(x+d)(y+\deltay)-y(x+\deltax - d)}
$$

$$
\therefore y2 = 2d*\frac{y^2+y\deltay}{xy+yd+x\deltay+d\deltay -xy-y\deltax + yd}
$$
$$
\therefore y2 = 2d*\frac{y^2+y\deltay}{2yd+x\deltay+d\deltay -y\deltax }
$$

Thus, the observed error in $y$ (i.e., $y2-y$) would be given by 

$$
\therefore y2-y = \frac{2d * y^2 +  2d y\deltay}{2yd+x\deltay+d\deltay -y\deltax }-y
$$

$$
\therefore y2-y =  \frac{2d * y^2 + 2d y \deltay - 2d*y^2- xy\deltay-yd\deltay+y^2\deltax}{2yd+x\deltay+d\deltay -y\deltax }
$$

$$
\therefore y2-y =  \frac{d y \deltay - xy\deltay+y^2\deltax}{2yd+x\deltay+d\deltay -y\deltax }
$$

Since $\deltax$ and $\deltay$ are small, the denominator can be approximated as $2yd$, to yield
$$
 y2-y\approx \frac{d  \deltay - x\deltay+y\deltax}{2d }
$$

The above formula gives the error in the $y$ coordinate. Next, we focus on the error in $x$ coordinate. From Section \ref{sec:computelocations}, $x2 = \frac{y2}{\tan(\alpha_1)} -d $. Thus, we have

$$
\therefore x2 = 2d*\frac{y^2+y\deltay}{2yd+x\deltay+d\deltay -y\deltax }*\frac{x+d}{y} -d 
$$

$$
\therefore x2 = 2d*\frac{(y+\deltay)(x+d)}{2yd+x\deltay+d\deltay -y\deltax } -d 
$$

$$
\therefore x2-x = 2d*\frac{(y+\deltay)(x+d)}{2yd+x\deltay+d\deltay -y\deltax} -(x+d) 
$$

$$
\therefore x2-x = (x+d)\left( \frac{2d( y+\deltay)}{2yd+x\deltay+d\deltay -y\deltax } -1\right) 
$$

\begin{footnotesize}
$$
\therefore x2-x = (x+d) \frac{2d y + 2d\deltay- 2yd -x\deltay - d \deltay + y \deltax}{2yd+x\deltay+d\deltay -y\deltax } 
$$

\end{footnotesize}

$$
\therefore x2-x = (x+d) \frac{ -x\deltay + d \deltay + y \deltax}{2yd+x\deltay+d\deltay -y\deltax } 
$$

Since $\deltax$ and $\deltay$ are very small, the denominator can be approximated as $2yd$, to yield
$$
 x2-x \approx (x+d) \frac{ -x\deltay + d \deltay + y \deltax}{2yd } 
$$

$$
\therefore x2-x \approx \frac{(x+d)}{y}\cdot \frac{ -x\deltay + d \deltay + y \deltax}{2d } 
$$

Since the total error is $\sqrt{(x2-x)^2+(y2-y)^2}$, 
we have, expected error ($\expectederrortwod$) is given by 
$
\expectederrortwod = |\frac{ -x\deltay + d \deltay + y \deltax}{2d}(\sqrt{\frac{(x+d)^2}{y^2} + 1})|
$

We consider a few special cases. (1) $\deltax=0$, and (2) $\deltay=0$.
In the first case, $\expectederrortwod$ is $|\frac{ -x\deltay + d \deltay}{2d}(\sqrt{\frac{(x+d)^2}{y^2} + 1})|$, which is equal to $|\frac{\deltay}{2d} (d-x)(\sqrt{\frac{(x+d)^2}{y^2} + 1})| $. Thus, the error is likely to highest when $y$ coordinate is low. We observe this in Figure \ref{fig:dy0.01} and \ref{fig:dy0.001}.
In the second case, the expected error is $|\frac{ y \deltax}{2d}(\sqrt{\frac{(x+d)^2}{y^2} + 1})|$, which is equal to $|\frac{ \deltax}{2d}(\sqrt{{(x+d)^2} + y^2})|$
. Hence, error is likely to be highest when $x$ coordinate is high. We observe this in Figure \ref{fig:dx0.01} and \ref{fig:dx0.001}.


\vspace*{-3mm}
\section{Simulation of Localization Error for 2D Camera}
\label{sec:2Dexperiments}
\vspace*{-3mm}

In this section, we focus on the location error when the cameras take pictures at different times. 
Specifically, we begin with a location $(x, y)$. We use the angles $\alpha_1$, $\alpha_2$, $\beta_1$, and $\beta_2$ identified by Camera 1 and 2. We use $\deltax$ and $\deltay$ to be errors introduced in the location. Recall that the values of $\deltax$ and $\deltay$ would depend upon the timing error and the rate of movement of the given point. Since the cumulative effect of the timing error and movement is captured by the actual movement, we directly focus on the actual movement rather than the individual parameter of timing error and rate of movement. 
Subsequently, we use $x1$ and $y1$ to determine $\alpha_2'$ and $\beta_2'$. Finally, we use $\alpha_1$, $\beta_1$, $\alpha_2'$ and $\beta_2'$ to determine $x2$ and $y2$.


Based on \cite{ndivega}, in this analysis, we let $d=25$. All units are in centimeters.  
Thus, the distance between the cameras is 50cm. 
Additionally, based on the operating range of NDI Vega Camera \cite{ndivega}, we consider the case where $x$ value is between $-70$ to $70$ and $y$ value is between $90$ to $240$. 

\begin{figure*}[!t]
    \centering
    \subfloat[]{
        \includegraphics[width=0.45\textwidth]{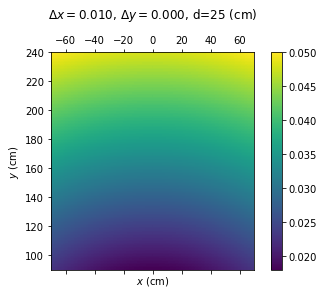}
        \label{fig:dx0.01}}
    \hfill
    \subfloat[]{
        \includegraphics[width=0.45\textwidth]{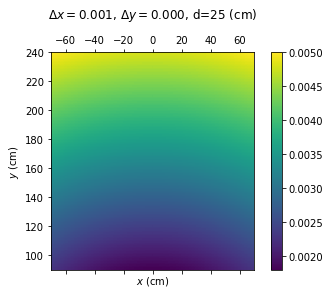}
        \label{fig:dx0.001}}
    \\
    \subfloat[]{
        \includegraphics[width=0.45\textwidth]{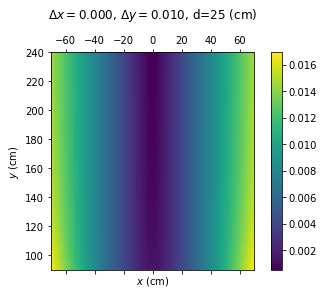}
        \label{fig:dy0.01}}
    \hfill
    \subfloat[]{
        \includegraphics[width=0.45\textwidth]{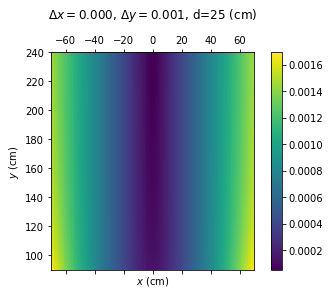}
        \label{fig:dy0.001}}
    \\
    \subfloat[]{
        \includegraphics[width=0.45\textwidth]{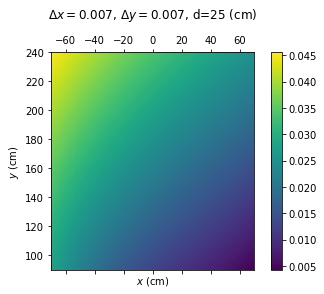}
        \label{fig:dx0.001dy0.001}}
    \hfill
    \subfloat[]{
        \includegraphics[width=0.45\textwidth]{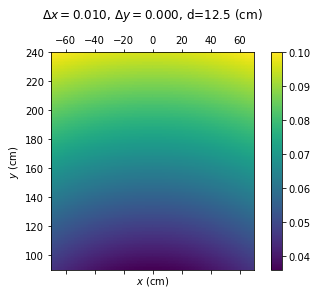}
        \label{fig:d=12.5}}
    \vspace*{-3mm}
    
    \caption{2D Location Error 
    }
    \label{fig:2Dlocationerror}
    \vspace*{-7mm}
\end{figure*}

\textbf{Simulation 1: Perturbation along $X$ axis: }
First, we consider the case where $\deltax=0.01$, i.e., due to timing error, there is an error of 0.01 cm in the $x$ coordinate but there is no error in the $y$ coordinate.
Under these constraints, the observed error is as given in  Figure \ref{fig:dx0.01}. 
This analysis is consistent with the approximate errors identified in Section \ref{sec:2dsymbolic}. From the analysis in Section \ref{sec:2dsymbolic}, when $\deltay=0$, the error is highest when $(x+d)^2+y^2$ is highest. In other words, when $x$ value is fixed, error grows with $y$ value. And, when $y$ value is fixed, the error is highest when $x+d$ is highest.


We find that the observed error via simulation is consistent with this as well. 
Specifically, we find that the error is lowest when $x$ and $y$ are small, i.e., the location is closest to the center of the coordinate system. As we move farther the error increases. We consider an error of $0.01$ to be expected. If the pictures taken by the cameras had an error of $0.01$ then having the same error in the final calculation is expected. However, as we can see from this table, the error can be substantially higher. For example when  $x=70,y=240$, the error is $0.05$ (i.e., about 5 times more than expected). Furthermore, this error would increase even further if the camera were to be used for a larger range of $x$ and $y$ values. For example, if the usage of camera usage was extended to $x=100,y=400$ then the error would be $0.082$ (8.2 times larger than the actual movement). 

\textbf{Simulation 2: Perturbation along $Y$ axis: }
Figure \ref{fig:dy0.01} considers the case where the error is caused in $y$ coordinate, i.e.,  $\deltax=0$, $\deltay=0.01$. In this case, we observe that errors are highest when you are farther from the $Y$ axis. In particular, when (the absolute value of) $x$ coordinate is high, we observe that the error is high as well. For example, for $x=-70,y=90$, the error is approximately $0.017$ (1.7 times the actual error $\deltay$). The errors are lower when you are farther from $X$ axis. For example, when $x=0, y=240$, the error reduces to approximately $0.002$. 
This analysis is also consistent with the results from Section \ref{sec:2dsymbolic}. Additionally, if camera usage were extended for lower values of $Y$ then the error would be even higher. For example, if the camera were to be used when $Y=40$, the error would increase to $0.048$ (4.8 times the actual error in $\deltay$).

\textbf{Simulation 3: Diagonal perturbation: }
Figure \ref{fig:dx0.001dy0.001} considers the case where the error is diagonal ($\deltax=\deltay=\frac{0.01}{\sqrt{2}}$). Here, we observe that the error is highest at $x=-70,y=240$ where the error is $0.046$ (4.6 times the actual movement of 0.01). 
We note that the results for $\deltax=\frac{0.01}{\sqrt{2}}$ and $\deltay=- \frac{0.01}{\sqrt{2}}$ are similar except that the maximum error occurs at $x=70,y=240$.
Furthermore, if camera usage were to be increased for a larger range, say $x=\pm 100$ and $y=400$, the error would increase to $0.073$ (7.3 times larger than the actual movement).  


\textbf{Simulation 4: Scaling $\deltax$ and $\deltay$:} We also consider the case where $\deltax=0.001$ or $\deltay=0.001$. The results are as shown in Figures \ref{fig:dx0.001} and \ref{fig:dy0.001}. These results show that the overall effect is the same; if the actual error is reduced to $\frac{1}{10}^{th}$ then the observed error also reduces by the same margin. This is also consistent with the prediction in Section \ref{sec:2dsymbolic}.

\textbf{Simulation 5: Scaling of $d$: }
In our previous analysis, we considered $d=25$ based on a specific camera in \cite{ndivega}. Next, we consider the effect of scaling $d$, e.g., to make the camera system more compact. Figure \ref{fig:d=12.5} considers this case when $\deltax=0.01, \deltay=0$. Here, we observe that the error increases to $0.1$ (10 times the actual error caused by the timing error). Note that this error is twice that of the case when $d=25$. This is also predicted by the analysis in Section \ref{sec:2dsymbolic}.

\vspace*{-3mm}
\section{Modeling 3D Navigation Camera System}
\label{sec:3Dproblem}
\vspace*{-3mm}

In this section, we consider the extension of the problem in Section \ref{sec:2Dproblem} to the 3D setting.
In the 3D setting, the problem is similar to that in Figure \ref{fig:cameradiagram} except for the addition of $Z$ axis which is perpendicular to the paper. The definition of $X$ and $Y$ axis remains the same. Hence, the cameras uses angles $\alpha_1$, $\beta_1$ and $\gamma_1$ (respectively $\alpha_2, \beta_2$ and $\gamma_2$) with $X$, $Y$ and $Z$ axis respectively. Subsequently, the system identifies the location $(x,y,z)$ of point $C1$ using these six angles. 

Similar to the 2D analysis, based on the specification of \cite{ndivega}, we consider the case where $X$ coordinate varies from $-70$ to $+70$, $Y$ coordinate varies from $90$ to $240$. In the 3D analysis the $Z$ coordinate varies from $-65$ to $+65$. 

One key difference between 2D and 3D is that in 2D,  $\alpha_1+\beta_1=\frac{\pi}{2}$. But in 3D, these angles are independent.

\vspace*{-3mm}

\section{Analysis of 3D Camera Problem}
\label{sec:3DAnalysis}

\vspace*{-3mm}

 The structure of this section is similar to that of Section \ref{sec:2Dproblem}. Specifically, in Section \ref{sec:3Dlocation}, we consider how the angles reported by the cameras can be used in locating point $C1$. Section \ref{sec:3Dangles} identifies the angles reported by the camera based on the location of point $C1$. Section \ref{sec:3dsymbolic} considers the symbolic analysis to determine location error whereas Section \ref{sec:3Dsimulation} considers the case via simulation. 

\subsection{Identifying Location of Point Based on Camera Angles}
\label{sec:3Dlocation}

In this section, we extend the analysis from Section \ref{sec:computelocations} for the 3D setting. 
%
Similar to the analysis in Section \ref{sec:computelocations}, we have\footnote{In this system of equations, we have five unknowns, $x,y,z,r_1$ and $r_2$ and we have 6 equations. In other words, the equations include redundant information. Note that this is not a concern when angles are reported from the same location. However, it will create certain concerns when the points under consideration differ.}

\begin{equation}
x = r_1 \cos(\alpha_1) - d = r_2 \cos(\alpha_2) + d \label{eq:x3d}
\end{equation}
\vspace*{-5mm}
\begin{equation}
y = r_1 \cos(\beta_1)  = r_2 \cos(\beta_2)  \label{eq:y3d}
\end{equation}
\vspace*{-5mm}
\begin{equation}
z = r_1 \cos(\gamma_1)  = r_2 \cos(\gamma_2)  \label{eq:z3d}
\end{equation}

Since Equations \ref{eq:x3d} and \ref{eq:y3d} are identical to Equations \ref{eq:xval} and \ref{eq:yval} respectively,  using the same analysis in Section \ref{sec:computelocations}, we have  
$$y = \frac{2d * \cos (\beta_1) * \cos(\beta_2)}{\cos(\beta_2)*\cos(\alpha_1) - \cos(\beta_1)*\cos(\alpha_2)}
$$

Furthermore, since Equation $\ref{eq:y3d}$ is similar to Equation \ref{eq:z3d}, we have 

$$  z = \frac{2d * \cos (\gamma_1) * \cos(\gamma_2)}{\cos(\gamma_2)*\cos(\alpha_1) - \cos(\gamma_1)*\cos(\alpha_2)}
$$


Note that we cannot use the same approach as in Section \ref{sec:2Dproblem} to compute the $x$ coordinate. This is due to the fact that $\alpha_1$ and $\beta_1$ are now independent.
Instead, we can use Equation \ref{eq:y3d} to find $r_1 = \frac{y}{\cos \beta_1}$ and use this in Equation \ref{eq:x3d}. It gives us

$$
x = \frac{y}{\cos \beta_1}* \cos(\alpha_1) - d 
$$

$$ \therefore
x = \frac{2d * \cos (\beta_1) * \cos(\beta_2)}{\cos(\beta_2)*\cos(\alpha_1) - \cos(\beta_1)*\cos(\alpha_2)}*\frac{\cos \alpha_1}{\cos \beta_1}*  - d 
$$

Note that this formula also works in 2D setting since $\cos(\beta_1) = \sin(\alpha_1)$ in 2D setting where $\alpha_1 + \beta_1 = \pi/2$. 

\subsection{Identifying Angles Based on Location}
\label{sec:3Dangles}

In this section, we consider the case where the point pictured by cameras is at location $(x, y, z)$. Then, we identify the corresponding angles provided by each camera. 
Identifying angles in 3D space can be done via the approach for computing angle between vectors.  The angle, $\theta$ between two vectors A and B is given by $\arccos(\frac{\vec{A}.\vec{B}}{|A||B|})$  

The $X$ (respectively $Y$ and $Z$) axis can be viewed as vector $\langle 1, 0, 0\rangle$ (respectively, $\langle 0, 1, 0 \rangle, \langle 0, 0, 1 \rangle$). The vector representing the line between the point $(x, y, z)$ and the location of Camera 1 (-d, 0, 0) (respectively, Camera 2 (+d, 0, 0) is $(x+d, y, z)$ (respectively $(x-d, y, z)$). Thus, we have 

$$
\alpha_1 = \arccos{\frac{x+d}{r_1}}, \ \ \ 
\beta_1 = \arccos{\frac{y}{r_1}}, \ \ \ 
\gamma_1 = \arccos{\frac{z}{r_1}}
$$

$$
\alpha_2 = \arccos{\frac{x-d}{r_2}}, \ \ \ 
\beta_2 = \arccos{\frac{y}{r_2}}, \ \ \ 
\gamma_2 = \arccos{\frac{z}{r_2}}
$$

where $r_1 = \sqrt{((x+d)^2 + y^2 + z^2)}$, and $r_2 = \sqrt{((x-d)^2 + y^2 + z^2)}$,

\subsection{Symbolic Analysis of Localization Error due to Timing Error}
\label{sec:3dsymbolic}

In this section, we extend the analysis from Section \ref{sec:2dsymbolic}. Specifically, we consider the case where the instrument held by the surgeon is at location $(x, y,z)$ when the picture is taken by Camera 1. However, the instrument was at location $(x1, y1, z1)$ when picture was taken by Camera 2, and $x1=x+\deltax$, $y1=y+\deltay$ and $z1 = z+\deltaz$. where $dx, \deltay$ and $\deltaz$ are small values. 

As noted in Section \ref{sec:3Dlocation}, the system of Equations contains a redundant equation. This leads to an inconsistency where the values of $x2, y2$ and $z2$ that satisfy Equations \ref{eq:x3d}, \ref{eq:y3d} and \ref{eq:z3d} does not exist. This is due to the fact that the corresponding vectors may not intersect. However, given that the perturbation is small, we can compute the value by any five of the equations. In our analysis, we use the same formula as derived in Section \ref{sec:3Dlocation} except that the values of $\alpha_2,\beta_2$ and $\gamma_2$ would be determined by the location $(x1, y1, z1)$. 
Thus, the computed value of $y2$ would be given by 

$$  y2 = \frac{2d * \cos (\beta_1) * \cos(\beta_2)}{\cos(\beta_2)*\cos(\alpha_1) - \cos(\beta_1)*\cos(\alpha_2)}
$$

$$  \therefore y2 = \frac{2d * \frac{y}{r1} * \frac{y+\deltay}{r2'}}{\frac{y +\deltay}{r2'}*\frac{x+d}{r1} - \frac{y}{r1}*\frac{x-d+dx}{r2'}}
$$

$$  \therefore y2 = \frac{2d * {y} * (y+\deltay)}
{(y +\deltay)*(x+d) - y ({x-d+dx})}
$$

Since this formula is identical to that in Section \ref{sec:2dsymbolic}, the error in the $y$ location (namely $y2-y$) would be given by the same formula in Section \ref{sec:2dsymbolic}. Thus, we have 
$\frac{d\deltay -x\deltay + y\deltax}{2d}$ 
Using the same analysis, we have 
$\frac{d\deltaz -x\deltaz + z\deltax}{2d}$ 

Finally, we also observe that the analysis in Section \ref{sec:2dsymbolic} is also valid for computing $x2-x$. In 2D, $\alpha_1+\beta_1= \pi/2$. This is not true in 3D analysis. However, in spite of that we have 
$\cos(\alpha_1)=\frac{x+d}{r1}$ and 
$\cos(\beta_1)=\frac{y}{r_1}$. Hence, $\frac{\cos(\alpha_1)}{\cos(\beta_1)}= \frac{x+d}{y}$. This is the exact formula used in Section \ref{sec:2dsymbolic} to find the error $x2-x$ in the $X$ coordinate.  Thus, we have
$x2-x \approx \frac{(x+d)}{y} \frac{ -x\deltay + d \deltay + y \deltax}{2d } 
$

We note that we have computed the error in each coordinate. The combined error would be $\sqrt{(x2-x)^2+(y2-y)^2+(z2-z)^2}$.

%


\begin{figure*}[!t]
    \centering
    \subfloat[]{
        \includegraphics[width=0.45\textwidth]{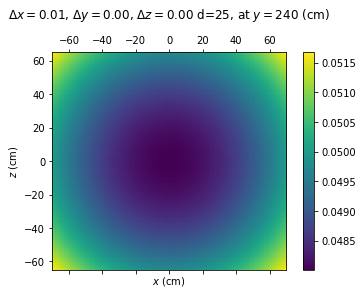}
        \label{fig:3Ddx0.1}}
    \hfill
    \subfloat[]{
        \includegraphics[width=0.45\textwidth]{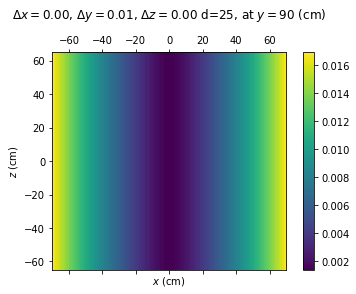}
        \label{fig:3Ddy0.01}}
    \hspace*{-10mm}\\
    \subfloat[]{
        \includegraphics[width=0.45\textwidth]{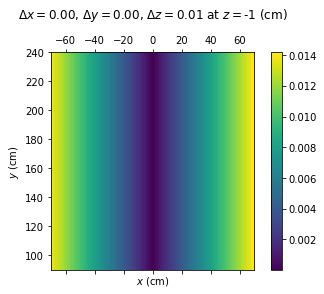}
        \label{fig:3Ddz0.01}}
    \hfill
    \subfloat[]{
        \includegraphics[width=0.45\textwidth]{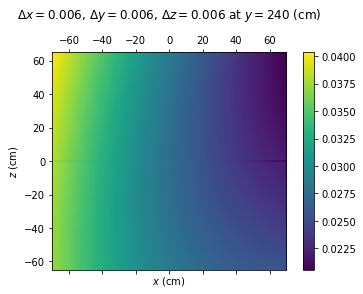}
        \label{fig:3Ddxdydz0.01-sqrt3}}
   \hspace*{-10mm} \\
    \subfloat[]{
        \includegraphics[width=0.45\textwidth]{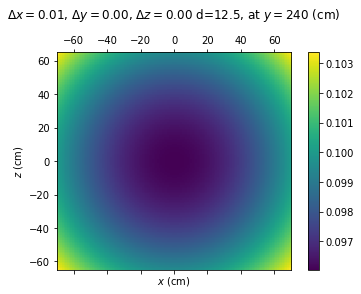}
        \label{fig:3Dd=12.5}}

    \caption{3D Location Error 
    }
    \label{fig:3Dlocationerror}
\end{figure*}

\vspace*{-3mm}

\section{Simulation of Localization Error for 3D Camera}
\label{sec:3Dsimulation}
\vspace*{-3mm}


In this section, we consider the localization error in the 3D Camera setting. Similar to the analysis in Section \ref{sec:2Dexperiments}, we consider $(\deltax=0.01,\deltay=0,\deltaz=0)$, 
$(\deltax=0,\deltay=0.01,\deltaz=0)$, 
$(\deltax=0,\deltay=0,\deltaz=0.01)$, and
$(\deltax=\deltay=\deltaz=\frac{0.01}{\sqrt{3}})$. We also consider the effect of changing $d$. 


From Figures  \ref{fig:dx0.01} and  \ref{fig:dx0.001}, we observe that when the error is along the $X$ axis, the error is highest when $Y$ value is high. Hence, we consider the error analysis when $Y$ values are high. Specifically, consider the analysis when $Y=240$ and vary $X$ from $-70$ to $70$ and $Z$ from $-65$ to $+65$. The results are as shown in Figure \ref{fig:3Ddx0.1}.
From this figure, we observe that the overall error is almost the same (but \textit{slightly higher}) as that in Figure \ref{fig:dx0.01}. In part, this is due to the fact that the overall error is $\sqrt{(x2-x)^2 + (y2-y)^2 + (z2-z)^2}$, and the contribution of $z$ is low. For example, at $x=70,y=240,z=65$, the location error along all axis is $(0.095, 0.48, 0.13)$. This implies the overall contribution due to the addition of the $Z$ coordinate is low.  The maximum error for this simulation was $0.052$ and one of the points where it happens is (70, 240, -65).

From Figure \ref{fig:dy0.01}, if the perturbation is along $Y$ axis then we find that the error is maximum when $Y$ value is low. Hence, we consider the case where $Y=90$ and vary $X$ from $-70$ to $+70$ and $Z$ from $-65$ to $+65$. From Figure \ref{fig:3Ddy0.01}, we observe that the location error is approximately $0.017$. One of the points where this error occurred was 
$(-70,90,-62)$.

Figure \ref{fig:3Ddz0.01} considers the case when the movement is along $Z$ axis. Here, the maximum error was $0.014$ and one of the points where it occurred was (70, 105, -1).
%
%
Figure \ref{fig:3Ddxdydz0.01-sqrt3} considers the case where the perturbation is diagonal $(\deltax=\deltay=\deltaz=\frac{0.01}{\sqrt{3}}\approx0.006$. Here the maximum error is $0.04$ and one of the locations where this error occurs is (-70, 240, 65). 
Finally, the effect of changing $d$ is similar to the 2D model. Specifically, from Figure \ref{fig:3Dd=12.5}, we find that if the value of $d$ is changed from 25 to $12.5$ the observed location in the error is $0.14$, twice the value compared with Figure \ref{fig:3Ddx0.1}.

\section{Related Work}
\label{sec:related}

Our work is orthogonal to the work in \cite{ptp,8969519} which focuses on building better time synchronization mechanism. The work in \cite{8325927} focuses on building a monitor to detect timing errors. We anticipate that in the problems considered here, while a monitor could be useful for long-term analysis to detect and correct problems, it is unlikely to assist in an online setting where the cameras are used to guide the surgeon. There is a substantial work (e.g., \cite{5340224}) to evaluate attacks on protocols such as NTP. However, these attacks are unlikely to occur in closed systems such as \cite{ndivega}.

There is substantial work  \cite{DBLP:journals/sensors/ZhangWQ21,DBLP:conf/cvpr/TakahashiMIK18,DBLP:conf/cvpr/AndrilukaPGS14} in the area of computer vision where pictures taken at different times are combined to form a composite picture. In this work, it is already known that the pictures were taken at different times. By contrast, we are focusing on the scenario where pictures are believed to be taken at the same time and the exact timing error is often unknown at runtime. 


\section{Conclusion}
\label{sec:concl}
\vspace*{-3mm}

In this paper, we analyzed the errors caused by timing errors in cyber-physical systems. Cyber-physical systems rely on some events happening simultaneously or periodically. However, a timing error could cause a slight delay between the events that are intended to be simultaneous. 

Our focus was on the surgical camera from \cite{ndivega} (which is  an instance of a navigational camera system). We find that the effect of the timing error depends upon the location of the device when the timing error occurs. Any timing error in a moving object would result in some localization error. Hence, our focus was on the question: If there is a timing error of $\epsilon$ and the object moves by a distance $\delta$ during that time, what is the effective error observed by the user. 

We find that in the operating region of the camera in \cite{ndivega}, the error observed by the user is magnified by as much as five times. In other words, if there is a timing error of $\epsilon$ and the object moves by a distance within time $\epsilon$ then the observed error is as high as $5\delta$.
As expected, the observed error is proportional to $\delta$. 

We also find the location of the object and direction of the movement affect the observed error. For example, we find that if the movement is along the $X$ axis then the observed error is maximum when the instrument is far away from the cameras. On the other hand, if the movement is along the $Y$ axis then the observed error is maximum when the instrument is closer to the camera. 

We also identify how the distance between two cameras affects the observed error. In particular, if the distance between cameras is reduced by half (e.g., to make the system more compact) then the observed error doubles.

Our work shows that errors caused by timing errors can have a substantial impact on the operating range of these systems. For example, even with a timing error of $15 \mu s$ at speed of $40 cm/s$, the localization error would be $0.003cm$ which is 10\% of the accuracy provided by \cite{ndivega}, i.e., even small timing errors can consume a substantial part of the safety margin. 
This error is even higher if the operating range of the camera is increased or the camera system is changed to be more compact.  Furthermore, timing errors are caused by various reasons including clock skew, unanticipated interrupts, different execution times along different branches of the program, etc. Additionally, the problems caused by timing errors increase as the speed increase or systems become more compact (e.g., $d$ decreases).

There are several future applications of this work. Our analysis shows that errors caused by timing errors in safety-critical systems need to be evaluated carefully. As systems become larger they tend to utilize multiple subsystems increasing the likelihood of timing errors. Furthermore, the effect of timing error may be magnified in the error observed by the user. 

\bibliographystyle{plain}
\bibliography{camerabib}
\end{document}